\documentclass[letterpaper, 10 pt, conference]{ieeeconf}  % Comment this line out if you need a4paper

\IEEEoverridecommandlockouts                              % This command is only needed if 
\usepackage{amsmath} % assumes amsmath package installed
\usepackage{amssymb}  % assumes amsmath package installed
\usepackage{adjustbox}
\usepackage{array}
\usepackage{multirow}
\usepackage{booktabs}
\usepackage{cite}
\usepackage{graphics}
\usepackage{stfloats}

\usepackage[draft,inline,nomargin]{fixme}

\title{\LARGE \bf
Learning Robot Activities from First-Person Human Videos\\
Using Convolutional Future Regression
}

\author{Jangwon Lee and Michael S. Ryoo% <-this % stops a space
\thanks{School of Informatics and Computing, Indiana University, Bloomington, IN 47408, USA.
\{leejang, mryoo\}@indiana.edu}
}
\begin{document}

\maketitle
\thispagestyle{empty}
\pagestyle{empty}

%%%%%%%%%%%%%%%%%%%%%%%%%%%%%%%%%%%%%%%%%%%%%%%%%%%%%%%%%%%%%%%%%%%%%%%%%%%%%%%%
\begin{abstract}
We design a new approach that allows robot learning of new activities from unlabeled human example videos. Given videos of humans executing the same activity from a human's viewpoint (i.e., first-person videos), our objective is to make the robot learn the temporal structure of the activity as its future regression network, and learn to transfer such model for its own motor execution. We present a new deep learning model: We extend the state-of-the-art convolutional object detection network for the representation/estimation of human hands in training videos, and newly introduce the concept of using a fully convolutional network to regress (i.e., predict) the intermediate scene representation corresponding to the future frame (e.g., 1-2 seconds later). Combining these allows direct prediction of future locations of human hands and objects, which enables the robot to infer the motor control plan using our manipulation network. We experimentally confirm that our approach makes learning of robot activities from unlabeled human interaction videos possible, and demonstrate that our robot is able to execute the learned collaborative activities in real-time directly based on its camera input.

%Providing first-person human videos to the robot is as if we are providing the robot ‘visual memory’ of itself performing the activities previously.

\end{abstract}

%%%%%%%%%%%%%%%%%%%%%%%%%%%%%%%%%%%%%%%%%%%%%%%%%%%%%%%%%%%%%%%%%%%%%%%%%%%%%%%%

\section{Introduction}

One of the important abilities of humans (and animals) is
that they are able to learn new activities
and their motor controls from others' behaviors.
When a person watches others performing an activity,
he/she not only learns to visually predict future consequences of the motion during the activity
but also learns how to execute the activity himself/herself.

Recently, approaches taking advantage of ``deep learning'' for robot manipulation
have been gaining an increasing amount of attention,
directly learning motor control policies given visual inputs (i.e., images and videos) \cite{levine2016end}.
The use of convolutional neural networks (CNNs) have been particularly successful,
since they are able to jointly learn image features optimized for the task
based on their training data.
Because of such ability, new models incorporating convolutional and recurrent neural networks (i.e., CNNs and RNNs)
is likely to become a major trend in robotics, just like what already happened in computer vision and machine learning.

However, although these deep learning oriented approaches showed very promising results on learning video prediction \cite{finn2016unsupervised}
and actual motor control policy \cite{levine2016end},
they have been limited to relatively simple actions such as object grasping and pushing.
This is because a large amount of `robot' data is necessary for the direct training of these CNNs and RNNs with millions of parameters.
A large number of samples of humans (or the robot itself) motor controlling the robot is necessary for generating training data \cite{levine2016end},
and this is a limiting aspect particularly when we want to teach a robot new (i.e., previously unseen) activities.

In this paper, we present a new CNN-based approach
that enables robot learning of its activities from `human' example videos.
Human activity videos can be attractive training resources because it does not require any hardware
or professional software for teaching robots,
even though it might create other difficulties like transferring learned human-based models to the actual robots.
Given videos of humans executing the same activity from a human's viewpoint (i.e., first-person videos),
our objective is to make the robot learn the temporal structure of the activity as its future regression network,
and learn to transfer such model for its own motor execution.
The idea is that a human's first-person video and the video a humanoid robot is expected to obtain during its activity execution should be very similar.
Providing first-person human videos to the robot is as if we are providing the robot `visual memory' of itself performing the activities previously.
This enables the robot to directly learn what visual observation it is expected to see during the correct execution of the activity and how it will change from its viewpoint.

%There have been (non-CNN-based) previous works on robot activity learning from human videos \cite{lee2013syntactic},
There have been previous works on robot activity learning from human videos \cite{lee2013syntactic, yang2015robot},
extending the previous concept of `robot learning from demonstration' \cite{argall2009survey} which was mostly done with direct motor control data.
However, these works focused on learning grammar representations of human activities,
modeling human activities as a sequence of atomic actions (e.g., grasping).
These approaches were limited in the aspect that activities were always represented in terms of pre-defined set of atomic actions,
and the users had to teach the robot how to recognize those atomic actions from human activity videos by providing labeled training data (i.e., supervised learning).
This prevented the robot learning of activities from scratch,
and was also limited in that human had to define new atomic actions when a new activity is added.
Furthermore, since it was not trainable in an end-to-end fashion,
the robot has to somehow figure out how to execute those atomic actions, which was usually done by hand-coding the motion.

\begin{figure*}[t]
 \centering
    \includegraphics[width=\textwidth]{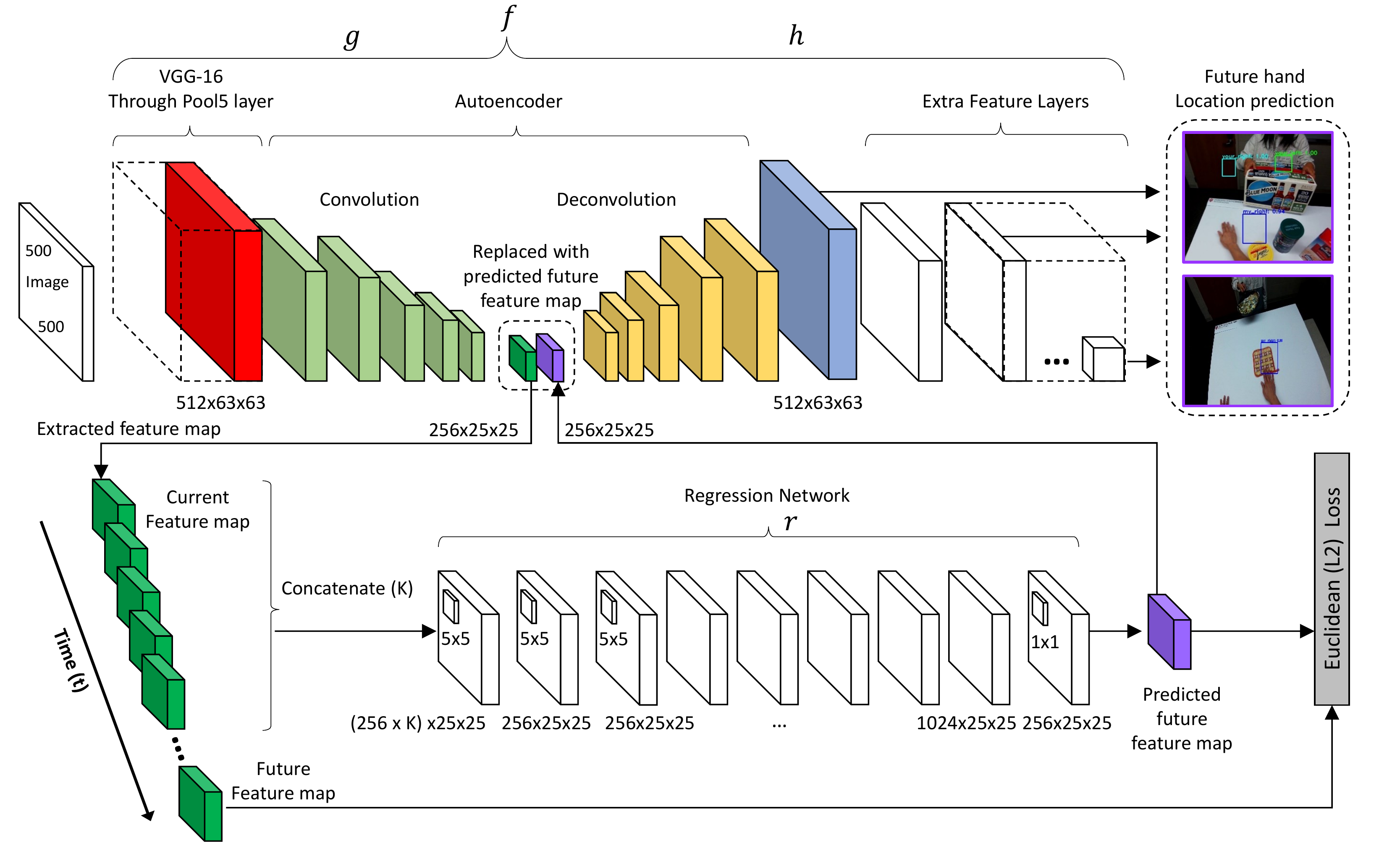}
 \caption{Overview of our perception component:
          Our perception component consists of two fully convolutional neural networks:
          The first network is an extended version of the state-of-the-art convolutional object detection network
          (SSD \cite{ssd}) for the representation of human hands and estimation of the bounding boxes (top).
          The second network is a future regression network to regress (i.e., predict)
          the intermediate scene representation corresponding to the future frame.
          This network does not require activity labels or hand/object labels in videos for its training.}
 \label{fig:our_network}
\end{figure*}

We introduce a new robot activity learning model using a fully convolutional network for future representation regression.
We extend the state-of-the-art convolutional object detection network (SSD \cite{ssd}) for the representation of human hand-object information in a video frame,
and newly introduce the concept of using a fully convolutional network to regress (i.e., predict) how such intermediate scene representation will change in the future frame (e.g., 1-2 seconds later).
Combining these allows direct and explicit prediction of future hand locations (Figure \ref{fig:our_network}), which then allows the robot to infer the motor control plan.
That is, not only feature-level prediction of future representations (similar to \cite{vondrick2015anticipating})
but also semantic-level prediction of explicit future hand locations of humans and robots during the learned activity is being jointly performed in our new network.
Such future hand prediction results are used by our manipulation network that learns mapping of the 2-D hand locations in the image coordinate to the actual motor control.

Our activity learning is an unsupervised approach in the aspect that it does not require activity labels or hand/object labels in the activity videos.
It does require hand-annotated training data for the learning of its hand representation network,
but there already exists public datasets for this purpose and it does not require any labels for its future regression network.
The future regression network is learned without supervision by capturing changes in our hand-based representations in the training videos.
In addition, importantly, all our networks were designed to function in real-time for the actual robot operation, and we show such capability with our experiments in this paper.

\section{Related work}

\paragraph{Robot learning from humans}
There have been a considerable amount of previous efforts on robot learning from demonstration (LfD) \cite{argall2009survey, billard2008robot, gupta2016learning}. Since it enables robots automatically learn a new task from demonstration by non-robotics expert,
LfD is very important in robotics. However, there are limitations since most of these approaches focused on making robots learn motor control polices from human data, which usually was in the form of direct control sequences obtained with actual robots or simulation softwares \cite{thomaz2009learning}. Moreover, it often requires a knowledge about all primitive actions for teaching high-level tasks \cite{mulling2013learning}.

There also have been previous works on robot activity learning from visual data \cite{lee2013syntactic,yang2015robot,shu2016learning},
extending the previous concept of LfD. These works focused on learning grammar representations of human activities from conventional third-person videos (i.e., videos usually taken with static cameras watching the actors), modeling human activities as a sequence of atomic actions (e.g., grasping).
Having a grammar representation composed of atomic actions allows transfer of human activity structure to robots, and the robot replication of human activities was possible usually with hand-coded motion transfer from human atomic actions to robot atomic actions. However, activity learning was generally done in a fully supervised fashion with human annotations in these approaches, and they assumed very reliable estimation of semantic features from videos such human hands and human body skeletons. \cite{koppula2014} studied an approach to directly learn object manipulation trajectories from human videos, but it was limited to one-robot-one-object scenarios unlike our approach focusing on very general human-robot collaboration scenarios (e.g., human-object-robot interactions).

\paragraph{Video prediction}

Our approach in this paper is to generate proper robot behaviors (particularly for human-robot collaboration) by predicting `future' visual representation. The idea is that such representation leads to the estimation of future positions of objects and hands of humans and robots. Visual prediction is one of the core components of our perception system.

There have been previous works on the
prediction of future frames from the computer vision community
\cite{vondrick2015anticipating,walker2014patch,lotter2016deep}. However, there has been very limited attempt on applying such future predictions for robotics systems, since these approaches in general requires more components for interpreting predicted representation to generate robot actions. In the above works, no robot manipulation was actually attempted. There exists a recent robotics work that attempted applying visual prediction
for generating robot control actions \cite{finn2016deep}.
This study shows the potential in applying visual prediction
for a robotic manipulation task; it enables transferring the visual perception
to robot manipulation component for generating motor control commands
without any additional components to interpret the recognition results.
However, this requires a huge amount of training data using actual physical robots to make the robot learn activities, and thus is limited when the robot needs to learn many new activities.

\paragraph{First-person videos}

First-person videos, also called egocentric videos, are the videos taken from the actor's own viewpoint. Recognition of human/robot activities from such first-person videos has been actively studied particularly in the past 5 years, including recognition of human actions from wearable cameras \cite{kitani11,fathi11,ramanan12,ryoo15} and human-robot interactions from robot cameras \cite{ryoo15hri,ryoo16icra}. However, these focused on building discriminative video classifiers, and the attempt to learn `executable' representations of human activities or their transfer to robots have been very limited.

The main contribution of this paper is in enabling robot activity learning from human interaction videos using our newly proposed convolutional future regression. We believe this is the first work to present a deep learning-based (i.e., entirely CNN-based) method for learning human-robot interactions from human-human videos. We also believe this is the first paper to take advantage of human `first-person videos' for the robot activity learning.
\section{Approach}

\subsection{System Overview}

Given a sequence of current frames,
our goal is to (i) predict future hand locations and all interactive objects
in front of the robot, then to (ii) generate robot control commands for moving
robot's hands to the predicted hand locations.
We employ two components for achieving the goal.
The first component is a perception component that consists of two fully convolutional neural networks:
(1) an extended version of the Single Shot MultiBox Detector (SSD) \cite{ssd} to create a hand-based scene representation and estimate bounding boxes, and (2) a future regression network to model how such intermediate scene representation (should) change in future frames.
The second component is a manipulation component that maps 2-D hand locations in the image coordinate to the actual motor control using fully connected layers.

The key idea of our approach is that the proposed perception component allows prediction of future (1-2 seconds later) hand locations given current video input from a camera.
Such future prediction can be learned based on humans' first-person activity videos by using them as training data, with the assumption that the robot camera has a similar viewpoint with the human first-person videos.
This allows the robot to directly predict its ideal future hand locations during the activity, inferring how the hand should move if the activity were to be executed successfully.
Next, the manipulation component generates actual robot control commands to move the robot's hands to the predicted future locations.

%Since we use human's first-person activity videos as training data for teaching the robot and a video frame of the robot camera has similar viewpoint as the first-person videos, the robot can consider the future locations of a camera wearer's hands as desired positions of his/her hands.

\subsection{Perception Component}

Given a video frame $\mathbf{\hat{X}}_{t}$ at time $t$, the goal 
of our perception component is to predict the future
hand locations $\mathbf{\hat{Y}}_{t+\Delta}$.

\paragraph{Hand Representation Network}

%For achieving our goal,
We first construct a network for the hand-based representation of the image scene by extending the SSD object detection framework. We extended it by inserting
a fully convolutional auto-encoder having five convolutional layers followed by five deconvolutional layers for dimensionality reduction. This allows the approach to abstract an image (with hands and objects) into a lower dimensional intermediate representation.

All our convolutional/deconvolutional layers use 5$\times$5 kernels
and the number of filters for each convolutional layer are:
512, 256, 128, 64, 256. The green convolutional layers in Fig. \ref{fig:our_network} correspond to them.
After such convolutional layers, there are deconvolutional layers (yellow layers in Fig. \ref{fig:our_network}), each having the symmetric number of filters: 256, 64, 128, 256, 512.  
We do not use any pooling layer, and
instead use stride 2 for the last convolutional layer 
for the dimensionality reduction. We thus increase the number of filters 
for the last convolutional layer to compensate loss of information.

Let $f$ denote the hand representation network given an image at time $t$.
Then, this network can be considered as a combination of two sub functions, $f = g \circ h$:
\begin{equation} \label{eq:1}
\mathbf{\hat{Y}}_{t} = f(\mathbf{\hat{X}}_{t}) = h(\mathbf{\hat{F}}_{t}) = h(g(\mathbf{\hat{X}}_{t})),
\end{equation}
where a function $g:\mathbf{\hat{X}} \rightarrow \mathbf{\hat{F}}$
denotes a feature extractor (from an input video frame to encoder)
to get compressed intermediate visual representation (i.e., feature map) $\mathbf{\hat{F}}$,
and $h:\mathbf{\hat{F}} \rightarrow \mathbf{\hat{Y}}$
indicates a box estimator which uses the compressed representation as an input
for locating hand boxes at time $t$.
With the  above formulation, the network can predict hand locations $\mathbf{\hat{Y}}_{t}$ at time $t$ after the training.

\paragraph{Future Regression Network}

\begin{figure}[t]
 \centering
    \includegraphics[width=\linewidth]{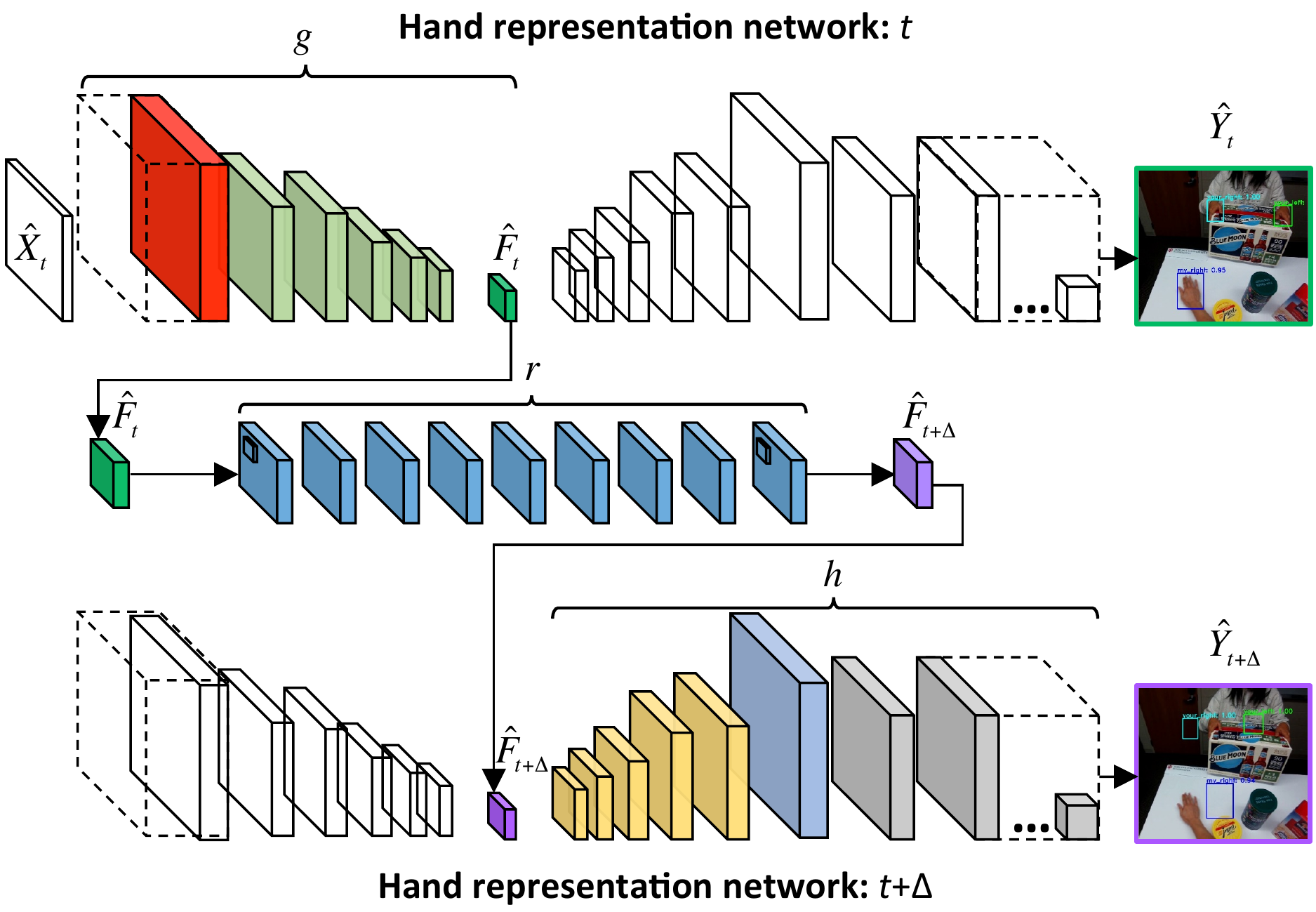}
 \caption{Data flow of our perception component during test phase.
          It enables predicting hands corresponding to the future frame. Only the colored layers are  used for the prediction in the test phase.}
 \label{fig:data_flow}
\end{figure}

Although the above hand representation network allows obtaining hand boxes in the `current' frame, our objective is to get the `future' hand locations $\mathbf{\hat{Y}}_{t+\Delta}$
instead of theirs current locations $\mathbf{\hat{Y}}_{t}$.

We formulate this problem as a regression problem. The main idea is that the intermediate representation of the hand representation network $\mathbf{\hat{F}}_{t}$ abstracts the hand-object information in the scene, and that we are able to take advantage of it to infer the future (intermediate) representation $\mathbf{\hat{F}}_{t +\Delta}$. Once such regression becomes possible, we can simply plug-in the predicted future representation $\mathbf{\hat{F}}_{t +\Delta}$ to the remaining part of the hand network (i.e., $h$) to obtain the final future hand prediction results. Therefore, we newly design a network for predicting the intermediate scene representation corresponding to the future frame
${\hat{F}}_{t +\Delta}$, as a fully convolutional future regression network: Fig. \ref{fig:data_flow}.

%We are visualizing the two applications of the hand detection network, one for the current frame and the other for the future frame.

Given a current scene representation $\mathbf{\hat{F}}_{t}$
from the hand network, our future regression network ($r$) predicts the future the intermediate scene representation $\mathbf{\hat{F}}_{t +\Delta}$:
\begin{equation} \label{eq:2}
\mathbf{\hat{F}}_{t +\Delta} = r_{w}(\mathbf{\hat{F}}_{t}).
\end{equation}
It has seven convolutional layers having 256 5$\times$5 kernels. In addition, it has a layer with 1024 13$\times$13 kernels followed by the last layer that has 256 1$\times$1 kernel. We trained the weights ($w$) of the regression network with unlabeled first-person human activity videos using the following loss function:
\begin{align} \label{eq:3}
w^* &= \operatorname*{arg\,min}_{w}\sum_{i,t}\| r_{w}(\mathbf{\hat{F}}^i_{t}) - \mathbf{\hat{F}}^i_{t +\Delta} \|_{2}^2 \nonumber \\
  &= \operatorname*{arg\,min}_{w}\sum_{i,t}\| r_{w}(g(\mathbf{\hat{X}}^i_{t})) - \mathbf{\hat{F}}^i_{t +\Delta} \|_{2}^2
\end{align}
where $\mathbf{\hat{X}}^i_{t}$ indicates a video frame at time $t$ from video $i$,
and $\mathbf{\hat{F}}^i_{t}$ represents a feature map at time $t$ from video $i$.

Our future regression network can use any intermediate scene representation
from any intermediate layers of the hand network,
but we use the one from auto-encoder due to its lower dimensionality.
Finally, the future scene representation $\mathbf{\hat{F}}_{t +\Delta}$
is fed into the hand network for estimating hand boxes corresponding to the future frame
to get future hand locations $\mathbf{\hat{Y}}_{t+\Delta}$. 
\begin{equation} \label{eq:4}
\mathbf{\hat{Y}}_{t+\Delta} = h(\mathbf{\hat{F}}_{t+\Delta})
\end{equation}

Fig.~\ref{fig:data_flow} summarizes data flow of our perception component during testing phase.
Given a video frame $\mathbf{\hat{X}}_{t}$ at time $t$,
(1) we extract the intermediate scene representation $\mathbf{\hat{F}}_{t}$
using the feature extractor ($g$), and then (2) feed it into the future regression network ($r$)
to get future scene representation $\mathbf{\hat{F}}_{t +\Delta}$.
Next, (3) we feed $\mathbf{\hat{F}}_{t +\Delta}$ into the box estimator ($h$),
and finally obtain future position of hands $\mathbf{\hat{Y}}_{t+\Delta}$ at time $t$.
\begin{equation} \label{eq:5}
\mathbf{\hat{Y}}_{t+\Delta} = h(\mathbf{\hat{F}}_{t+\Delta}) = h(r(\mathbf{\hat{F}}_{t}))  = h(r(g(\mathbf{\hat{X}}_{t})))
\end{equation}
Furthermore, instead of using just a single frame (i.e., the current frame) for the future regression, we extend our network to take advantage of the previous $K$ frames to obtain $\mathbf{\hat{F}}_{t+\Delta}$ as illustrated in Fig. \ref{fig:our_network}:
\begin{equation}
\mathbf{\hat{Y}}_{t+\Delta} = h(r([g(\mathbf{\hat{X}}_{t}), ..., g(\mathbf{\hat{X}}_{t-(K-1)}) ])).
\end{equation}

The advantage of our formulation is that it allows us to predict future hand locations while considering the implicit activity and object context, even without explicit detection of objects in the scene. Our auto-encoder-based intermediate representation $\mathbf{\hat{F}}^i_{t}$ abstracts the scene configuration by internally representing what objects/hands are currently in the scene and where they are, and our fully convolutional future regressor takes advantage of it for the prediction.

\begin{figure*}[th]
 \centering
    \includegraphics[width=\textwidth]{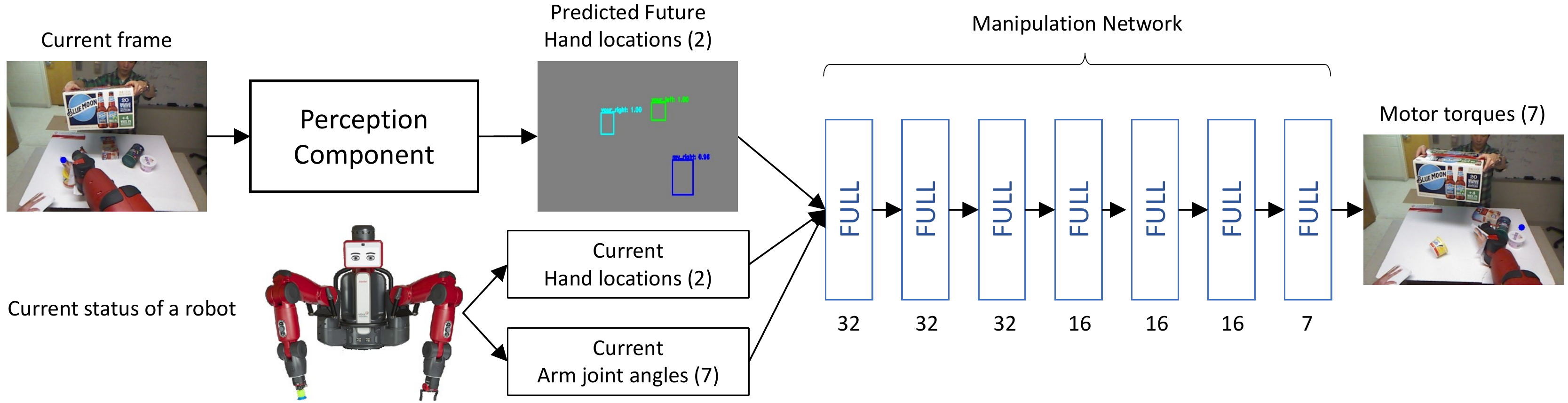}
 \caption{Robot manipulation component of our approach.
          It generates robot control commands given current robot joint state,
          current robot hand locations,
          and predicted future robot hand locations.}
 \label{fig:robot_network} 
\end{figure*}

\subsection{Manipulation Component}
\label{subsec:manipulation}

Although our perception component is able to predict future hand locations of humans in first-person human activity videos, it is insufficient for the robot manipulation. Here, we construct another regression network ($m$) for mapping the predicted 2-D human hand locations in the image coordinate to the actual motor control commands. The main %assumption is that a video frame from a robot's camera will have a similar viewpoint to our training data (first-person human activity videos), allowing us to take advantage of the learned model for the robot future hand prediction:
assumption is that a video frame from a robot's camera will have a similar viewpoint to our training data (first-person human videos), allowing us to take advantage of the learned model for the robot future hand prediction by assuming:
\begin{equation} \label{eq:6}
    \mathbf{\hat{Y}}_{\mathbf{R}t} \simeq \mathbf{\hat{Y}}_{t}
\end{equation}
where, $\mathbf{\hat{Y}}_{\mathbf{R}t}$ represents robot hand locations.

Our manipulation component ($m$) predicts
future robot joint states ($\mathbf{\hat{Z}}_{t +\Delta}$)
given current robot joint states ($\mathbf{\hat{Z}}_{t}$),
robot hand locations ($\mathbf{\hat{Y}}_{\mathbf{R}t}$),
and future hand locations ($\mathbf{\hat{Y}}_{\mathbf{R}t +\Delta}$) telling where the robot's hands should move to.
This network can be formulated with the below function:
\begin{equation} \label{eq:7}
\mathbf{\hat{Z}}_{t +\Delta} =
m_{\theta}(\mathbf{\hat{Z}}_{t},\mathbf{\hat{Y}}_{\mathbf{R}t},\mathbf{\hat{Y}}_{\mathbf{R}t +\Delta}).
\end{equation}

Our manipulation component consists of seven fully connected layers having the following number of hidden units for each layer:
32, 32, 32, 16, 16, 16, 7.
The weights ($\theta$) of this network can be obtained
by the same way that used for our perception networks:
\begin{equation} \label{eq:8}
\theta^* = \operatorname*{arg\,min}_{\theta}\sum_{j,t}
\| m_{\theta}(\mathbf{\hat{Z}}^j_{t},\mathbf{\hat{Y}}^j_{\mathbf{R}t},\mathbf{\hat{Y}}^j_{\mathbf{R}t +\Delta})
- \mathbf{\hat{Z}}^j_{t +\Delta} \|_{2}^2
\end{equation}
where $\mathbf{\hat{Z}}^j_{t}$ indicates robot joint states at time $t$ from training episode $j$,
and $\mathbf{\hat{Y}}^j_{\mathbf{R}t}$ represents robot hand locations at time $t$ from training episode $j$.
Fig.~\ref{fig:robot_network} shows our manipulation component for generating robot control commands.

The combination of our perception component and manipulation component provides a real-time robotics system that takes raw video frames as its input and generates motor control commands for its activity execution. Our manipulation component can be replaced with a standard Inverse Kinematics, but our neural network-based model generates more natural arm movements by considering the desired location of the robot's end-effectors as well as joint configuration sequences (i.e., unlabeled robot logs described in the next section).

\section{Experiments}

\subsection{Datasets} \label{sec:datasets}

Our approach consists of three different types of networks (within the two components), and we use three different types of datasets for training each model.

\textbf{EgoHands\cite{Bambach_2015_ICCV}:}
This is a public dataset containing 48 first-person videos of people interacting in
four types of activities
(playing cards, playing chess, solving a puzzle, and playing Jenga).
It has 4,800 frames with 15,053 ground-truth hand labels.
Here, we added 466 frames with 1,267 ground-truth annotations to the original dataset to cover more hand postures.
We use this dataset to learn our hand representation network, which is trained to locate hand boxes in a video frame.

\textbf{Unlabeled Human-Human Interaction Videos:}
We collected a total of 47 first-person videos of human-human collaboration scenarios,
with each video clip ranging from 4 to 10 seconds.
This dataset is a main dataset for teaching a new task to our robot.
It contains two types of tasks: 
(1) a person wearing the camera cleaning up all objects on a table 
as a partner (i.e., the other subject) approaches the table while holding a heavy box (to make a room for her/him to put the heavy box on the table),
and (2) a person wearing the camera pushing a trivet on a table toward to a partner
when he/she is approaching the table while holding a hot cooking pan.
These videos are unlabeled videos without any activity/hand annotation and we trained our convolutional regression network
using this dataset.

%for predicitng future visual representation.

\textbf{Unlabeled Robot Activity Log Files:}
We prepared this dataset to train our robot manipulation network.
It contains 50 robot log files. Each log has the robot's hand positions ($u, v$) in an image plane and the robot's corresponding joint angles at time $t$. We recorded these log files by making a human operator move the robot arms (i.e., the human grabbed the robot arms and moved them). We obtained such robot joint configuration sequences while moving the robot to cover possible arm motion during general human-robot interaction tasks. Here, we assume that the robot is supposed to operate in a similar environment during the test phase. Note that this was not recorded under the interaction scenario (i.e., just the robot itself was moving), and no annotation regarding the activity or motion was provided.
We used a Baxter research robot for recording these files
and the Baxter has seven degrees-of-freedom arm: the file contains 9 variables for each arm.
In order to estimate the robot's hand position in the image plane,
we projected the 3-D positions of the Baxter's grippers into the image plane (based on camera calibration)
and recorded the projected ($u, v$) positions with 7 joint angles at 30 Hz.

\subsection{Baselines} \label{sec:baselines}

In order to provide quantitative comparisons,
we compared our perception component
with four different baselines: \textbf{(i) Hand-crafted representation}
uses a hand-crafted state representation based on explicit object and hand detection. It encodes relative distances
between all interactive objects in our two scenarios, and uses it to predict the future hand location using neural network-based regression.
More specifically, it detects objects using KAZE features \cite{alcantarilla2012kaze}
and hands using CNN based hand detector in \cite{Bambach_2015_ICCV},
then computes relative distances between all objects and hands
for building the state representation which is a 20 dimensional vector.
Then, we built a new network which has five fully connected layers trained using the state representations
on the same interaction dataset we use.
\textbf{(ii) Hands only} uses hand locations for the future regression.
It predicts future hand locations solely based on current hand locations
without considering any other visual representations.
In order to train this baseline model,
we extracted hand locations from all frames of the interaction videos using our hand representation network,
then made log files to store detected hand locations in each frame and their frame numbers.
After this, we trained another neural network model for the future hand location prediction using the log files, which has seven fully connected layers with the same number of hidden units as our robot manipulation network.
\textbf{(iii) SSD with future $\text{annotations}^{1}$}
is a baseline that uses the original SSD model \cite{ssd}
trained based on EgoHands dataset.
Instead of training the model to infer the current hand locations given the input frame, we fine-tuned this model on EgoHands dataset
after changing annotations of the dataset to have ``future'' locations of hands
instead of making it to use current hand locations.
We also used additionally 466 frames for this fine-tuning since the original EgoHands dataset was insufficient (too many repetitive hand movements) for this training.
\textbf{(iv) SSD with future $\text{annotations}^{2}$} is a baseline also using the original SSD model,
but we trained this model from scratch.
This time we changed all annotations of the EgoHands dataset,
then trained the model.
After that we fine-tuned the model as the same way
that used for the ``SSD with future $\text{annotations}^{1}$'' baseline.

\subsection{Evaluation of our future hand prediction} \label{sec:predction_results}

\begin{table}[t]
\centering
\caption{Evaluation of future hand prediction}
\begin{adjustbox}{max width=0.5\textwidth}
\begin{tabular}{ b{3.33cm} | l l l }
  \toprule
  \multirow{2}{*}{Method} & \multicolumn{3}{c}{Evaluation }\\
  \cmidrule{2-4}
   & Precision & Recall & F-measure \\
  \hline
  Hand-crafted representation & 0.30 $\pm$ 0.37 & 0.15 $\pm$ 0.19 & 0.20 $\pm$ 0.25 \\
  Hands only & 4.78 $\pm$ 3.70 & 5.06 $\pm$ 4.06 & 4.87 $\pm$ 3.81 \\
  SSD with future $\text{annotations}^{1}$ & 27.53 $\pm$ 23.36 & 9.09 $\pm$ 8.96 & 13.23 $\pm$ 12.62 \\
  SSD with future $\text{annotations}^{2}$ & 29.21 $\pm$ 19.16 & 7.92 $\pm$ 6.45 & 12.10 $\pm$ 9.42 \\
  \hline
  Deep Regressor (ours): K=1 & 27.04 $\pm$ 16.50 & 21.71 $\pm$ 14.71 & 23.45 $\pm$ 14.99 \\
  Deep Regressor (ours): K=5 & 29.97 $\pm$ 15.37 & 23.89 $\pm$ 16.45 & 25.40 $\pm$ 15.51 \\
  Deep Regressor (ours): K=10 & \textbf{36.58} $\pm$ 16.91 & \textbf{28.78} $\pm$ 17.96 & \textbf{30.90} $\pm$ 17.02 \\
  \bottomrule
\end{tabular}
\end{adjustbox}
 
\label{table:prediction}
\end{table}

We first evaluated the perception component of our approach in terms of precision, recall, and F-measure, and compared them against the above baselines. In the first evaluation, we made our approach to predict bounding boxes of human hands in the future frame given the current image frame. We measured the ``intersection over union'' ratio between areas of each predicted box and ground truth (future) hand locations.
Only when the ratio was greater than 0.5, the predicted box was accepted as a true positive. In this experiment, we randomly split the set of our Human-Human Interaction Videos into the training and testing sets, so 32 videos were used for training sets and remaining 15 videos were used for testing sets in a total of 47 videos.

%\mryoo{We need to describe more about the evaluation measures. We need to mention that we are predicting the hand bounding boxes in the `future' frame, and then are measuring intersection over union between such predicted boxes and ground truths. Only when the IoU is greater than 0.5, we accept the predicted box as a true positive. We also need to mention one more measure we are using, which is the mean pixel distance. Mention that we measure the distance only when both the ground truth hand and the predictions are present in the same scene.}

Table~\ref{table:prediction} shows
quantitative results of our future hand prediction.
Here, the plus-minus sign ($\pm$) indicates standard deviation and \textit{K} represents number of frames we used as an input
for our regression network. Our $\Delta$ was 30 frames (i.e., 1 sec).
We are able to clearly observe that our approach significantly outperforms all the baselines, including the state-of-the-art object detector SSD modified for the hand prediction.
Our proposed network with \textit{K} = 10 yielded
the best performance in terms of all three metrics,
at about 30.9 score in F-measure. The best performance we can get with SSD was only 13.23.

% need to add how we compute these score
% overlapratio > 0.5 can be used for threshold
% for considering true positive cases..

\begin{table}[t]
\centering
\caption{Mean pixel distance between ground truth and predicted positions of all hands}
\begin{adjustbox}{max width=0.505\textwidth}
\begin{tabular}{ b{4cm} | l }
  \toprule
  Method & Mean Pixel Distance\\
  \hline
  Hand-crafted representation & 143.85 $\pm$ 48.77 \\
  Hands only & 247.88 $\pm$ 121.94 \\
  SSD with future $\text{annotations}^{1}$ & 58.58 $\pm$ 36.76 \\
  SSD with future $\text{annotations}^{2}$ & 79.95 $\pm$ 102.07 \\
  \hline
  Deep Regressor (ours): K=1 & 51.31 $\pm$ 39.10 \\
  Deep Regressor (ours): K=5 & 51.41 $\pm$ 38.46 \\
  Deep Regressor (ours): K=10 & \textbf{46.66} $\pm$ 36.92 \\
  \bottomrule
\end{tabular}
\end{adjustbox}
 
\label{table:mean_distance}
\end{table}

\begin{table}[t]
\centering
\caption{Mean pixel distance between ground truth and predicted position of right hand}
\begin{adjustbox}{max width=0.505\textwidth}
\begin{tabular}{ b{4cm} | l }
  \toprule
  Method & Mean Pixel Distance\\
  \hline
  Hand-crafted representation & 121.48 $\pm$ 87.36 \\
  Hands only & 264.52 $\pm$ 148.15 \\
  SSD with future $\text{annotations}^{1}$ & 48.63 $\pm$ 39.04 \\
  SSD with future $\text{annotations}^{2}$ & 71.36 $\pm$ 104.18 \\
  \hline
  Deep Regressor (ours): K=1 & 40.08 $\pm$ 32.72 \\
  Deep Regressor (ours): K=5 & 40.46 $\pm$ 39.52 \\
  Deep Regressor (ours): K=10 & \textbf{36.78} $\pm$ 36.70 \\
  \bottomrule
\end{tabular}
\end{adjustbox}
 
\label{table:mean_distance_right}
\end{table}

In our second evaluation, we measured mean pixel distance between ground truth locations and the predicted positions of hands. The size of the image plane was 1280*720. We measured this mean pixel distance only when both the ground truths and the predictions are present in the same frame. Table~\ref{table:mean_distance} shows the mean pixel distance errors for all four types of hands (my left, my right, your left, and your right). Once more, we can confirm that our approaches greatly outperform the performance of all the baselines. The overall average distance was a bit high due to changes in human hand shapes and their variations, but they were sufficient in terms of generating robot motion.

\begin{figure*}[th]
 \centering
    \includegraphics[width=0.98\textwidth]{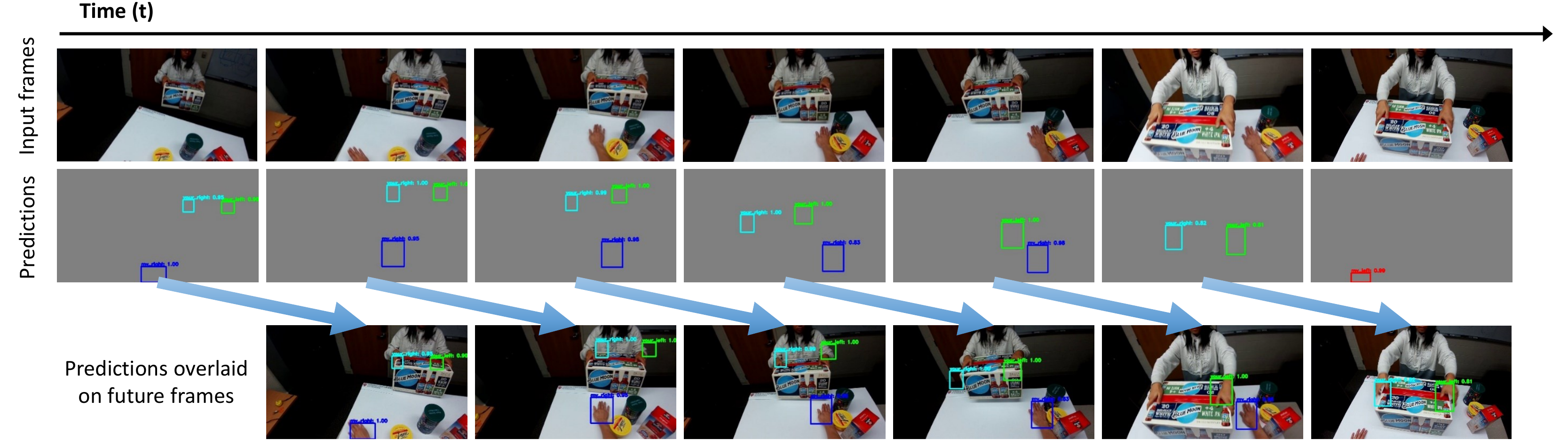}
    \includegraphics[width=0.98\textwidth]{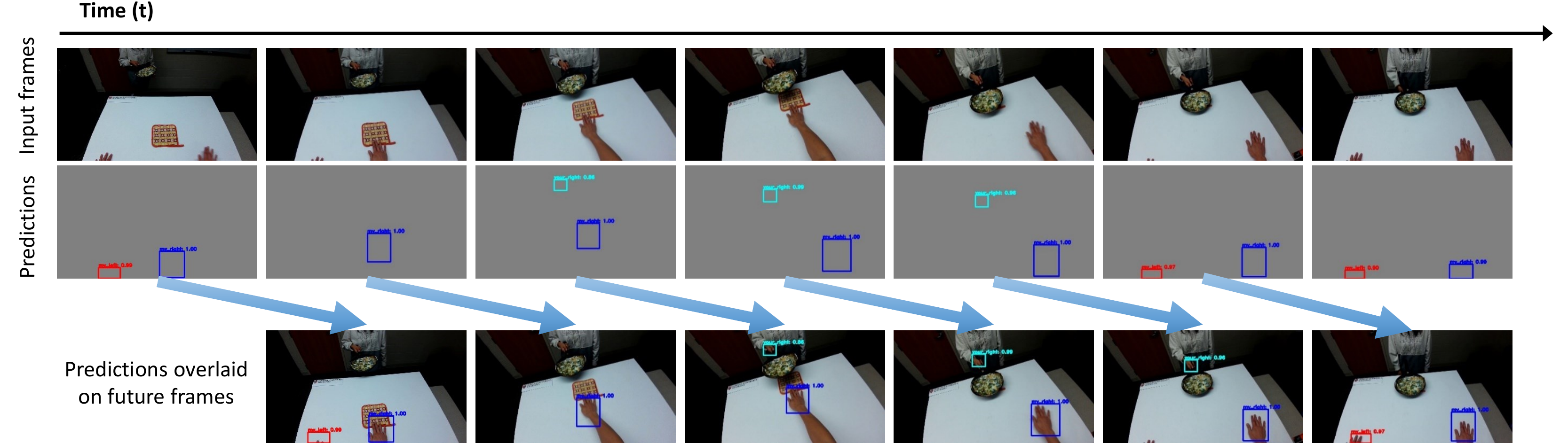}
 \caption{Two examples of our visual prediction. The first example is the activity of clearing the table, and the second example is the activity of pushing the trivet toward the person holding a cooking pan.
          The first row shows the input frames
          and the second row shows our future hand prediction results. In the third row, we overlaid our predictions on ``future'' frames. Red boxes correspond to the predicted `my left hand' locations, blue boxes correspond to `my right hand', green boxes correspond to the opponent's left hand, and the cyan boxes correspond to the opponent's right hand.
          The frames were captured every one second.}
 \label{fig:experiments}
\end{figure*}

We also compared accuracies of these methods while only considering  my right hand predictions, since position of my right hand is more important for a robot manipulation than locations of other types of hands. This is because, in our test scenarios, the robot's activities are very focused on its right hand motion. Table~\ref{table:mean_distance_right} shows mean pixel distance between ground truth and predicted position of `my right hand'. We can see that performances of our approaches are superior to all the baselines.
Examples of our visual predictions results are illustrated in Fig.~\ref{fig:experiments}.

\subsection{Real-time robot experiments} \label{sec:manipulation_results}

%Finally, we conducted a user study experiment evaluating the success of the robot activities performed based on our proposed approach.
% I just thought a user study experiment sound duplicated two words..
Finally, we conducted a user study to evaluate the success level of robot activities performed based on our proposed approach, with human subjects.
%We recruited twelve human subject participants, five undergraduate and seven graduate students,
A total of 12 participants (5 undergraduate and 7 graduate students) were recruited from the campus,
and were asked to perform one of the two activities (clearing the table for a partner and preparing a trivet for a cooking pan) together with our robot.
After such interactions, the participants were asked to complete a questionnaire about the robot behaviors for each task.
The questionnaire had two statements (one statement for each activity)
with scales from 1 (totally do not agree) to 5 (totally agree) to express their impression on the robot behaviors:
``I think the robot cleared the table to make a space for me.''
for the task 1 and ``I think the robot passed a trivet closer to me so that I can put the cooking pan on it.'' for the task 2.

%after giving them a brief introduction about this study.

%\mryoo{We need to talk about the scoring.}
%The input of the base control network is current 2D hand locations (image plane) and output of the network is seven joint angles corresponding to the 2D hand locations.

%\mryoo{What exactly is the `base control' network? I am getting confused. Please explain in more detail. What's its input and what's its output?}

In addition to our approach (i.e., our perception component + manipulation component), we designed and implemented the following three baselines and compared their quantitative results:
\textbf{(i) Base SSD + Base control} uses the baseline SSD with future $\text{annotations}^{1}$ as a perception component and the base manipulation network trained using the same robot activity log files. This base control network direct maps current hand locations in the image plane to current seven joint angles for each robot arm, without the $\mathbf{\hat{Z}}_{t}$ term in Eq. \ref{eq:7}.
%using the same network that uses 7 fully connected layers
% the main difference is ours use current joint angles as input, but base network does not use it, it just uses 2d hand postion
%so mapping is (2) variables (2d hand locations) to seven joint angles
\textbf{(ii) Base SSD + Our control} uses SSD with future $\text{annotations}^{1}$
as a perception component and our manipulation component (from Section \ref{subsec:manipulation}) to generate motor commands.
\textbf{(iii) Our perception + Base control} used our perception component
to predict future hand locations and the base control network for manipulation. In all these cases, the final control of our robot arm is performed by taking advantage of the Baxter API by providing the estimated future joint angle configuration.

\begin{figure*}[th]
 \centering
    \includegraphics[width=0.98\textwidth]{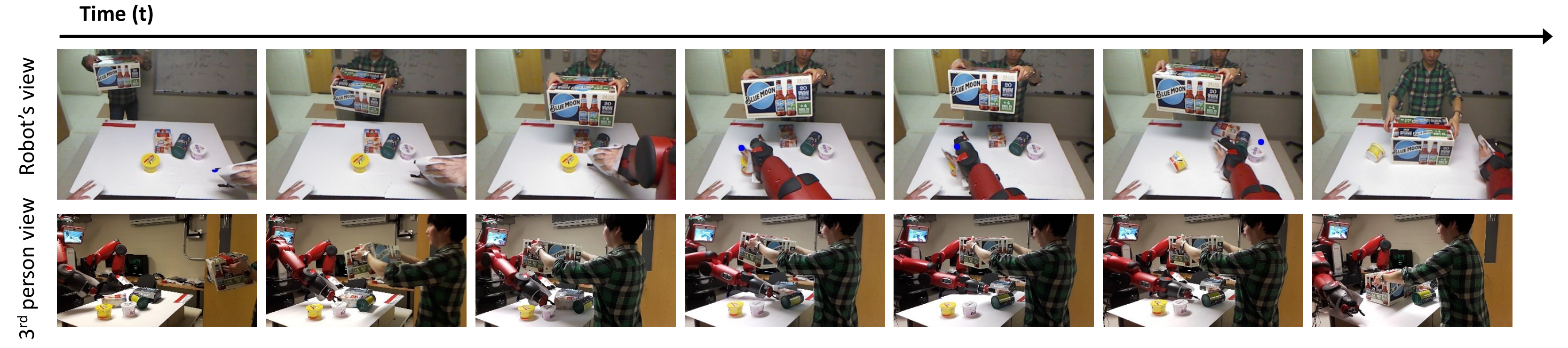}
    \includegraphics[width=0.98\textwidth]{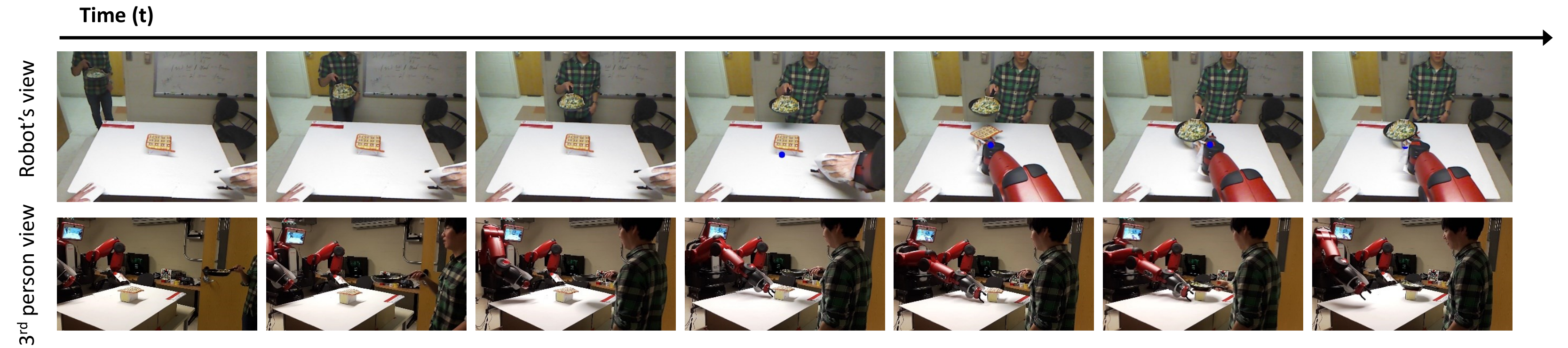}
 \caption{Qualitative results of our real-time robot experiments. Similar to Fig. \ref{fig:experiments}, there are two examples: clearing the table, and pushing the trivet toward the person.
 In each example, the first row shows the exact frames used as inputs to our robot (taken from a robot camera), and the second row shows the robot and the human from a 3rd person viewpoint. The frames were captured every one second.}
 \label{fig:robot_experiments}
\end{figure*}

As a result, each participant interacted with the robot total of 8 times in a random order.
Table~\ref{table:user_study} shows the results. The results indicate that our participants evaluated the robot with our approach performed better on both tasks.
We received a higher average score of 3.29 compared to all the baselines (1.72, 1.92, and 2.29) from the participants.
Examples of our real-time robot experiments with human subjects are illustrated in Fig.~\ref{fig:robot_experiments}.

Our method operates in slow real-time with our unoptimized C++ code. It takes $\sim$100 ms per frame using one Nvidia Pascal Titan X GPU, and we were able to conduct real-time human-robot collaboration experiments using it.

%\mryoo{Could you talk more about this mode?} 
% Well joint position control mode is just an of APIs from Baxter (rethink robotics).
% There are the below descriptions
%joint position control mode is the fundamental, basic control mode for Baxter arm motion. In position control mode, we specify joint angles at which we want the joints to achieve. Typically this will be consist of seven values, a commanded position for each of the seven joints, resulting in a full description of the arm configuration.
%that is just sending the all joint angles of arm than Baxter moves to achieve the joint angles.

%Our Z from mainplation network is seven joint angels, so I just push that joint angles to the controller.

\begin{table}[t]
\centering
%\caption{Experimental results evaluating the success level of our human-robot collaboration}
\caption{The success level of our human-robot collaboration}
\begin{adjustbox}{max width=0.489\textwidth}
\begin{tabular}{ b{3.4cm} | l l | l}
  \toprule
  Method & Task 1 & Task 2 & Average\\
  \hline
  Base SSD + Base control & 1.25 $\pm$ 0.43 & 2.21 $\pm$ 1.41 & 1.72 $\pm$ 0.92 \\
  Base SSD + Our control  & 1.5 $\pm$ 0.96 & 2.33 $\pm$ 1.60 & 1.92 $\pm$ 1.28 \\
  Our perception + Base control & 2.33 $\pm$ 1.18 & 2.25 $\pm$ 1.36 & 2.29 $\pm$ 1.27 \\
  Ours & \textbf{3.17} $\pm$ 1.40 & \textbf{3.42} $\pm$ 1.61 & \textbf{3.29} $\pm$ 1.50 \\
  \bottomrule
\end{tabular}
\end{adjustbox}
\label{table:user_study}
\end{table}

\section{Conclusion}
In this paper, we proposed a new robot activity learning model using a fully convolutional network for future representation regression. The main idea was to make the robot learn the temporal structure of a human activity as its future regression network, and learn to transfer such model for its own motor execution using our manipulation network. We show that our approach enables the robot to infer the motor control commands based on the prediction of future human hand locations in real-time. The experimental results confirm that our approach not only predicts the future locations of human/robot hands more reliably, but also is able to make robots execute the activities based on predictions. The paper focuses on robot learning of location-based hand movements (i.e., translations and natural rotations), and handling more dynamic hand posture changes remains as one of our future challenges.

%\section*{Acknowledgement}
{\flushleft\textbf{Acknowledgement:} This work was supported by the Army Research Laboratory under Cooperative Agreement Number W911NF-10-2-0016.}

\bibliographystyle{IEEEtran}
\bibliography{references}

\begin{thebibliography}{10}
\providecommand{\url}[1]{#1}
\csname url@samestyle\endcsname
\providecommand{\newblock}{\relax}
\providecommand{\bibinfo}[2]{#2}
\providecommand{\BIBentrySTDinterwordspacing}{\spaceskip=0pt\relax}
\providecommand{\BIBentryALTinterwordstretchfactor}{4}
\providecommand{\BIBentryALTinterwordspacing}{\spaceskip=\fontdimen2\font plus
\BIBentryALTinterwordstretchfactor\fontdimen3\font minus
  \fontdimen4\font\relax}
\providecommand{\BIBforeignlanguage}[2]{{%
\expandafter\ifx\csname l@#1\endcsname\relax
\typeout{** WARNING: IEEEtran.bst: No hyphenation pattern has been}%
\typeout{** loaded for the language `#1'. Using the pattern for}%
\typeout{** the default language instead.}%
\else
\language=\csname l@#1\endcsname
\fi
#2}}
\providecommand{\BIBdecl}{\relax}
\BIBdecl

\bibitem{levine2016end}
S.~Levine, C.~Finn, T.~Darrell, and P.~Abbeel, ``End-to-end training of deep
  visuomotor policies,'' \emph{Journal of Machine Learning Research}, 2016.

\bibitem{finn2016unsupervised}
C.~Finn, I.~Goodfellow, and S.~Levine, ``Unsupervised learning for physical
  interaction through video prediction,'' in \emph{Advances In Neural
  Information Processing Systems (NIPS)}, 2016.

\bibitem{lee2013syntactic}
K.~Lee, Y.~Su, T.-K. Kim, and Y.~Demiris, ``A syntactic approach to robot
  imitation learning using probabilistic activity grammars,'' \emph{Robotics
  and Autonomous Systems}, 2013.

\bibitem{yang2015robot}
Y.~Yang, Y.~Li, C.~Ferm{\"u}ller, and Y.~Aloimonos, ``Robot learning
  manipulation action plans by" watching" unconstrained videos from the world
  wide web.'' in \emph{AAAI}, 2015.

\bibitem{argall2009survey}
B.~D. Argall, S.~Chernova, M.~Veloso, and B.~Browning, ``A survey of robot
  learning from demonstration,'' \emph{Robotics and Autonomous Systems}, 2009.

\bibitem{ssd}
W.~Liu, D.~Anguelov, D.~Erhan, C.~Szegedy, S.~Reed, C.~Fu, and A.~Berg,
  ``{SSD}: Single shot multibox detector,'' in \emph{European Conference on
  Computer Vision (ECCV)}, 2016.

\bibitem{vondrick2015anticipating}
C.~Vondrick, H.~Pirsiavash, and A.~Torralba, ``Anticipating visual
  representations with unlabeled video,'' in \emph{IEEE Conference on Computer
  Vision and Pattern Recognition (CVPR)}, 2016.

\bibitem{billard2008robot}
A.~Billard, S.~Calinon, R.~Dillmann, and S.~Schaal, ``Robot programming by
  demonstration,'' in \emph{Springer handbook of robotics}, 2008.

\bibitem{gupta2016learning}
A.~Gupta, C.~Eppner, S.~Levine, and P.~Abbeel, ``Learning dexterous
  manipulation for a soft robotic hand from human demonstrations,'' in
  \emph{IEEE/RSJ International Conference on Intelligent Robots and Systems
  (IROS)}, 2016.

\bibitem{thomaz2009learning}
A.~L. Thomaz and M.~Cakmak, ``Learning about objects with human teachers,'' in
  \emph{ACM/IEEE International Conference on Human-Robot Interaction (HRI)},
  2009.

\bibitem{mulling2013learning}
K.~M{\"u}lling, J.~Kober, O.~Kroemer, and J.~Peters, ``Learning to select and
  generalize striking movements in robot table tennis,'' \emph{The
  International Journal of Robotics Research}, 2013.

\bibitem{shu2016learning}
T.~Shu, M.~S. Ryoo, and S.-C. Zhu, ``Learning social affordance for human-robot
  interaction,'' in \emph{International Joint Conference on Artificial
  Intelligence (IJCAI)}, 2016.

\bibitem{koppula2014}
H.~Koppula and A.~Saxena, ``Physically-grounded spatio-temporal object
  affordances,'' in \emph{European Conference on Computer Vision (ECCV)}, 2014.

\bibitem{walker2014patch}
J.~Walker, A.~Gupta, and M.~Hebert, ``Patch to the future: Unsupervised visual
  prediction,'' in \emph{IEEE Conference on Computer Vision and Pattern
  Recognition (CVPR)}, 2014.

\bibitem{lotter2016deep}
W.~Lotter, G.~Kreiman, and D.~Cox, ``Deep predictive coding networks for video
  prediction and unsupervised learning,'' \emph{arXiv preprint
  arXiv:1605.08104}, 2016.

\bibitem{finn2016deep}
C.~Finn and S.~Levine, ``Deep visual foresight for planning robot motion,'' in
  \emph{IEEE International Conference on Robotics and Automation (ICRA)}, 2017.

\bibitem{kitani11}
K.~M. Kitani, T.~Okabe, Y.~Sato, and A.~Sugimoto, ``Fast unsupervised
  ego-action learning for first-person sports videos,'' in \emph{IEEE
  Conference on Computer Vision and Pattern Recognition (CVPR)}, 2011.

\bibitem{fathi11}
A.~Fathi, A.~Farhadi, and J.~M. Rehg, ``Understanding egocentric activities,''
  in \emph{International Conference on Computer Vision (ICCV)}, 2011.

\bibitem{ramanan12}
H.~Pirsiavash and D.~Ramanan, ``Detecting activities of daily living in
  first-person camera views,'' in \emph{IEEE Conference on Computer Vision and
  Pattern Recognition (CVPR)}, 2012.

\bibitem{ryoo15}
M.~S. Ryoo, B.~Rothrock, and L.~Matthies, ``Pooled motion features for
  first-person videos,'' in \emph{IEEE Conference on Computer Vision and
  Pattern Recognition (CVPR)}, 2015.

\bibitem{ryoo15hri}
M.~S. Ryoo, T.~J. Fuchs, L.~Xia, J.~K. Aggarwal, and L.~Matthies,
  ``Robot-centric activity prediction from first-person videos: What will they
  do to me?'' in \emph{ACM/IEEE International Conference on Human-Robot
  Interaction (HRI)}, 2015.

\bibitem{ryoo16icra}
I.~Gori, J.~K. Aggarwal, L.~Matthies, and M.~S. Ryoo, ``Multi-type activity
  recognition in robot-centric scenarios,'' \emph{IEEE Robotics and Automation
  Letters (RA-L)}, 2016.

\bibitem{Bambach_2015_ICCV}
S.~Bambach, S.~Lee, D.~J. Crandall, and C.~Yu, ``Lending a hand: Detecting
  hands and recognizing activities in complex egocentric interactions,'' in
  \emph{IEEE International Conference on Computer Vision (ICCV)}, 2015.

\bibitem{alcantarilla2012kaze}
P.~F. Alcantarilla, A.~Bartoli, and A.~J. Davison, ``Kaze features,'' in
  \emph{European Conference on Computer Vision (ECCV)}, 2012.

\end{thebibliography}

\end{document}